\documentclass[10pt,twocolumn,letterpaper]{article}

\usepackage{cvpr}              

\definecolor{cvprblue}{rgb}{0.21,0.49,0.74}
\usepackage[pagebackref,breaklinks,colorlinks,allcolors=cvprblue]{hyperref}

\usepackage{graphicx}
\usepackage{booktabs}
\usepackage{multirow}
\usepackage{float}
\usepackage{dsfont}
\usepackage{amsfonts}
\usepackage{amssymb}
\usepackage{amsmath}
\usepackage{arydshln}
\usepackage{colortbl}
\usepackage[linesnumbered,ruled,vlined]{algorithm2e}
\usepackage{makecell}
\usepackage{appendix}
\usepackage[accsupp]{axessibility}

\SetCommentSty{mycommfont}
\DeclareMathAlphabet{\mathsfit}{\encodingdefault}{\sfdefault}{m}{sl}
\SetMathAlphabet{\mathsfit}{bold}{\encodingdefault}{\sfdefault}{bx}{n}

\newcommand{\abb}[0]{MAD }

\def\paperID{3} 
\def\confName{CVPR}
\def\confYear{2025}

\title{MAD: Makeup All-in-One with Cross-Domain Diffusion Model}

\author{Bo-Kai Ruan, Hong-Han Shuai\\
National Yang Ming Chiao Tung University\\
{\tt\small \{bkruan.ee11,hhshaui\}@nycu.edu.tw}\\\\
{Project Page: \url{https://basiclab.github.io/MAD}}
}

\begin{document}
 
\twocolumn[{%
\renewcommand\twocolumn[1][]{#1}%
\maketitle
\begin{center}
    \vspace{-15pt}
    \centering
    \captionsetup{type=figure}
    \includegraphics[width=\linewidth]{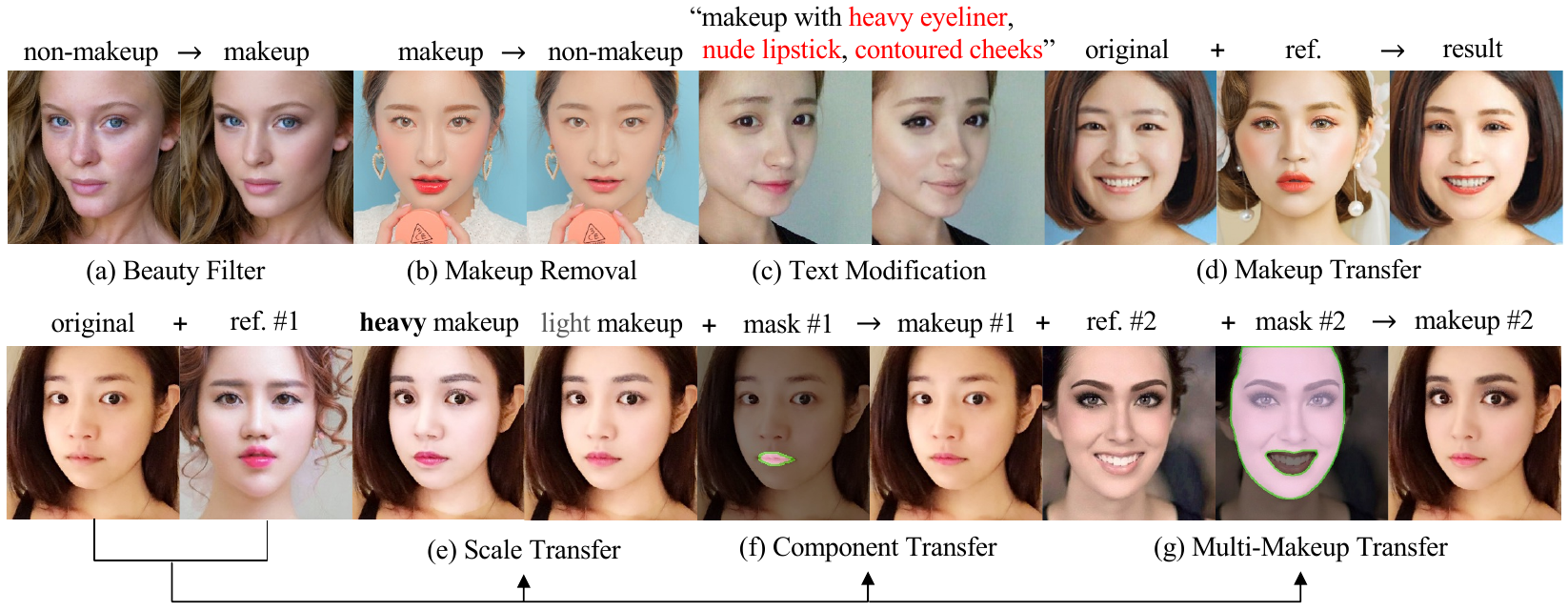}
    \captionof{figure}{The proposed MAD framework achieves a range of applications, including (a) beauty filter, (b) makeup removal, (c) text modification, (d) single-makeup transfer, (e) scale transfer, (f) component transfer, and (g) multi-makeup transfer.}
    \label{fig:teaser}
\end{center}%
}]

\begin{abstract}
\vspace*{-15pt}\\
Existing makeup techniques often require designing multiple models to handle different inputs and align features across domains for different makeup tasks, \textit{e.g.}, beauty filter, makeup transfer, and makeup removal, leading to increased complexity. Another limitation is the absence of text-guided makeup try-on, which is more user-friendly without needing reference images. In this study, we make the first attempt to use a single model for various makeup tasks. Specifically, we formulate different makeup tasks as cross-domain translations and leverage a cross-domain diffusion model to accomplish all tasks. Unlike existing methods that rely on separate encoder-decoder configurations or cycle-based mechanisms, we propose using different domain embeddings to facilitate domain control. This allows for seamless domain switching by merely changing embeddings with a single model, thereby reducing the reliance on additional modules for different tasks. Moreover, to support precise text-to-makeup applications, we introduce the MT-Text dataset by extending the MT dataset with textual annotations, advancing the practicality of makeup technologies.
\end{abstract}    
\vspace{-10pt}
\section{Introduction} \label{sec:intro}

Various makeup tasks have increasingly been conceptualized as generative challenges~\cite{zhu2017unpaired,chen2019beautyglow,liu2021psgan++,xiang2022ramgan,yan2023beautyrec}. For instance, in makeup transfer and removal, the CycleGAN framework~\cite{zhu2017unpaired} has emerged as a seminal approach, enabling the transfer and removal of makeup styles without paired training data. Despite its widespread adoption, CycleGAN and similar techniques face obstacles such as imprecise pose alignment~\cite{liu2021psgan++, xiang2022ramgan}, limitations in objective function design~\cite{yang2022elegant}, and difficulties in achieving accurate makeup correspondence~\cite{sun2022ssat}. To address these issues, various innovative solutions have been proposed, including advancements in pose alignment techniques~\cite{liu2021psgan++, xiang2022ramgan, deng2021spatially}, refinements of objective functions beyond basic histogram matching~\cite{yang2022elegant}, and methods ensuring precise facial semantic correspondence~\cite{sun2022ssat}. These enhancements aim to achieve more accurate makeup placement, thereby overcoming previous limitations and advancing state-of-the-art results.

However, despite these advancements, two major issues persist. \textbf{First}, current methodologies often require the use of \textit{multiple models} to perform diverse makeup-related tasks, introducing additional model complexity. This complexity arises primarily from difficulties in integrating features from disparate domains. Consequently, intricate model designs and complex training objectives are required, complicating the entire training process. For instance, many existing approaches resort to creating additional modules or employing multiple encoders and decoders to process and reconstruct features from different domains~\cite{xiang2022ramgan,yan2023beautyrec,sun2022ssat}. \textbf{Second}, although text-to-makeup transfer offers a user-friendly interface—for example, allowing users to try on makeup via voice commands converted through speech-to-text modules—existing approaches struggle to perform this task due to the \textit{lack of textual annotations}. This limitation is particularly acute in scenarios where textual descriptions are the sole source of reference, significantly restricting the versatility and applicability of text-to-makeup methods.

\begin{figure}[tbp]
 \centering
  \includegraphics[width=\linewidth]{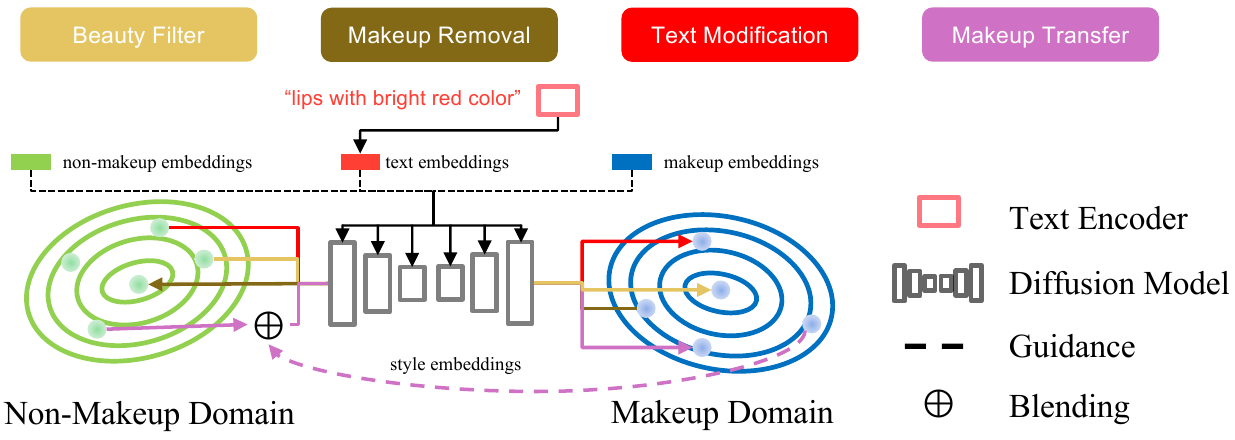}
  \caption{Each makeup task is modeled as a translation problem controlled by domain embeddings.}
  \label{fig:all-in-one-vis}
  \vspace{-10pt}
\end{figure}

To address the challenge, we propose a simple yet powerful model to perform a wide array of tasks, including (a) beauty filter, (b) makeup removal, (c) text modification, and various forms of makeup transfer, such as (d) single makeup transfer, (e) scale transfer, (f) component transfer, and (g) multi-makeup transfer, as illustrated in Fig.~\ref{fig:teaser}. The key observation is that these tasks inherently involve translating between domains with and without makeup. For example, beauty filter and makeup removal require navigation between these domains, while makeup transfer and text-driven modifications also translate from non-makeup to makeup but incorporate additional guidance, \textit{e.g.}, style references, or textual descriptions, into the process. Since the diffusion models have proved to be effective in cross-domain latent space manipulation~\cite{wu2023latent,hertz2023prompttoprompt,parmar2023zero} and text-to-image generation~\cite{rombach2022high}, we introduce the MAD (\textbf{M}akeup \textbf{A}ll-in-one with cross-domain \textbf{D}iffusion model). Fig.~\ref{fig:all-in-one-vis} illustrates the MAD framework built on the domain translation approach that separates the makeup and non-makeup domains and allows for the seamless execution of various makeup-related tasks through a cross-domain translation mechanism. 

Specifically, the foundational tasks from (a) to (d) are addressed through the concept of domain translation, while tasks (e) to (g) are considered subtasks of makeup transfer. For domain translation, we adopt the idea from CycleDiffusion~\cite{wu2023latent} and elegantly define the embedding for each domain. As shown in Fig.~\ref{fig:all-in-one-vis}, we define ``non-makeup'', ``makeup'', ``text'', and ``style'' embeddings and use only a single diffusion model. These embeddings are used as the conditional input to the diffusion model to serve as the domain controller. For example, the beauty filter and makeup removal tasks use the makeup and non-makeup embeddings, respectively, to either beautify an image or remove styles. These embeddings are learnable, initialized randomly, and refined through training with the diffusion model. Finally, to enhance text-based modification capabilities and address the second challenge, we have developed the MT-Text dataset for text-to-makeup annotations. This initiative significantly expands the potential for text-driven makeup transformations. The annotation process initiates with a request to GPT-4 Vision (GPT-4V)~\cite{achiam2023gpt} to delineate makeup styles for various facial components.

Our contributions can be summarized in three points:
\begin{itemize}
    \item To the best of our knowledge, we present the first one-for-all diffusion-based framework for a wide array of makeup tasks by modeling each as a domain translation problem.
    \item We introduce component asynchronous masking to provide accurate styles by applying masks at different steps to preserve style information for small components.
    \item We provide the MT-Text dataset, which is augmented from the existing makeup dataset with text descriptions to improve the applicability of text modification.
\end{itemize}

\section{Background and Preliminaries} \label{sec:background}

\subsection{Makeup Transfer}

Makeup transfer is a pivotal subset of style transfer, focusing on either imparting a specific makeup style from a reference image onto a source face or effectively removing existing makeup. Prior to the advent of deep learning, traditional methods like facial warping—relying on the alignment of facial landmarks—dominates the field~\cite{xu2013automatic}. These early techniques often employ heuristic methods to separate color layers for makeup transfer but struggle with style diversity and frequently produce underwhelming results. With the emergence of deep learning, a new line of studies leveraged Generative Adversarial Networks (GANs) to enhance synthesis quality. For instance, BeautyGAN~\cite{li2018beautygan} introduces a novel GAN-based approach that marks a significant milestone. Subsequent developments incorporate enhancements such as pose alignment and region-level makeup application to improve transfer quality. Notably, PSGAN++~\cite{liu2021psgan++} and RamGAN~\cite{xiang2022ramgan} integrate attention mechanisms to refine the process, while SCGAN~\cite{deng2021spatially} partition facial components into spatial-invariant style codes.

Further innovations in facial alignment have been presented in models like CPM~\cite{nguyen2021lipstick}, which employs a UV representation space to align the face for makeup transfer before re-projecting it back into 2D space. SSAT~\cite{sun2022ssat} focuses on facial semantic correspondence to enhance the precision of makeup location and style transfer. Moreover, EleGANt~\cite{yang2022elegant} has supported flexible local makeup editing directly within the feature space, enhancing the customizability of the transfer process. In pursuit of a more compact model design, BeautyREC~\cite{yan2023beautyrec} develops an architecture with a reduced parameter size that still efficiently performs makeup transfer tasks. Despite these advances, most existing models have relied on separate modules to align features across different domains and are primarily focused on makeup transfer, which limits their utility in executing various makeup-related tasks simultaneously. Additionally, the absence of a text-to-makeup dataset further restricts their application in text-based modification scenarios.

To address these limitations, we conceptualize each task as a cross-domain challenge by proposing a cross-domain diffusion model to handle multiple makeup tasks. Furthermore, to foster future research in text-based makeup applications, we introduce a novel text-to-makeup dataset, named MT-Text, to the existing MT dataset~\cite{li2018beautygan}. This enhancement significantly expands the dataset’s utility for multi-modal makeup tasks, promoting richer interactions and broader applicability in generative makeup systems.

\subsection{Technical Preliminaries and Notations}
The diffusion model uses the Markov chain and Gaussian distribution as a sample space~\cite{ho2020denoising}. By progressively introducing Gaussian noise into the images, the images are transformed into Gaussian noise. The Gaussian distribution governs the data transition from time $0$ to $t$ by:
\begin{equation} \label{eq:forward}
q(x_t|x_0) \triangleq \mathcal{N}(x_t | \sqrt{\alpha_t}x_0, (1 - \alpha_t)I),
\end{equation}
where $x_t$ is the noisy output at time $t \in \{0, 1, \cdots, T\}$, and $\alpha_t$ regulates the noise schedule. The diffusion model $f$ aims to master the reverse process, eliminating noise to reconstruct data at any given time $t$, and can generate real data from noise. The objective, which involves maximizing the evidence lower bound, can be simplified to an L2 norm:
\begin{equation}\label{eq:loss}
\mathcal{L} =  \| f(x_t, t) - \epsilon_t \|^2_2,
\end{equation}
where $\epsilon_t$ is the noise added to $x_0$ to yield $x_t$. As described in~\cite{song2021denoising}, the denoising process using the trained model $f$ to derive $x_{t-1}$ from $x_t$ is:
\begin{equation}
\begin{aligned}\label{eq:ddim}
x_{t-1} = &\sqrt{\alpha_{t-1}}\left(\frac{x_t - \sqrt{1 - \alpha_t}f(x_t, t)}{\sqrt{\alpha_t}}\right) \\
&+ \sqrt{1 - \alpha_{t - 1} - \sigma_t^2} \cdot f(x_t, t) + \sigma_t\epsilon_t,
\end{aligned}
\end{equation}
where $\sigma_t=\gamma \sqrt{(1 - \alpha_{t-1})/(1-\alpha_t)}\sqrt{1-\alpha_t/\alpha_{t - 1}}$, with $\gamma \in [0, 1]$ controlling the randomness scale. For simplicity, let $\mu_f(x_t, t)$ denote the first two terms, representing the mean estimator for time $t - 1$ using $f$.

\section{MAD Framework}
\label{sec:cross-domain-diffusion}

In this section, we first introduce the fundamentals of our framework in Sec.~\ref{sec:embeddings} and the acceleration approach in Sec.~\ref{sec:last-k} and elaborate on how to apply our framework to different makeup tasks in Sec.~\ref{sec:makeup-tasks}.

\subsection{Cross-Domain Translation with Embeddings} \label{sec:embeddings}

We first introduce how we conceptualize all makeup-related tasks as domain translation problems using domain embeddings. This framework builds upon the principles outlined in CycleDiffusion~\cite{wu2023latent}, which suggests using a diffusion model as a DPM-encoder to encode the source domain information within a single domain and then employing this encoded data as noise to generate outputs with another diffusion model in the target domain. For example, encoding a dog image to generate a cat image requires using the encoded output from a dog-specific diffusion model as the input noise for a cat-specific diffusion model.

Our extension of this concept eliminates the necessity of training separate diffusion models for each domain by introducing domain embeddings as conditional inputs. This allows for the control of output domains from a single diffusion model, covering non-makeup, makeup, text, and style domains, each represented by a specific embedding. Crucially, this means that after initial training of a diffusion model and the domain embeddings, no further fine-tuning is required for individual tasks, significantly streamlining the process. This approach not only simplifies the workflow but also enhances the adaptability of the model across multiple domains without compromising on the quality of domain translation. Given a source domain embedding $l_{so}$, a source domain input $x_0$, and a target domain embedding $l_{ta}$, the encoding process initiates by deriving the latent code $z_t$ through a forward diffusion process:
\begin{align}\label{eq:encoding}
    z_t &= (x_{t - 1} - \mu_f(x_t, t, l_{so})) / \sigma_t, \\
    x_{t - 1} &\sim q(x_{t - 1} | x_t, x_0).
\end{align}
Subsequently, the model reconstructs the image in the target domain using reverse generation:
\begin{equation}\label{eq:recon}
    \hat{x}_{t-1} = \mu_f(\hat{x}_{t}, t, l_{ta}) + \sigma_t z_t.
\end{equation}

To facilitate partial editing and ensure that areas not targeted for modification (\textit{e.g.}, hair and background) remain unchanged, a mask can be applied to allow users to define specific regions for alteration while preserving others. Inspired by~\cite{lugmayr2022repaint}, this mask is directly incorporated during the generation process without additional modules. This plug-and-play method involves adding noise to the preserved sections and sampling noise for the areas undergoing change, where the model then denoises the known sections and regenerates only the modified areas. The generation stage is enhanced accordingly:
\begin{align}
    \label{eq:mask-sample}
    \hat{x}'_{t - 1} &= \mu_f(\hat{x}_t, t, l_{ta}) + \sigma_t z_t, \\ \label{eq:mask-up}
    \hat{x}_{t - 1} &= M^s \cdot x_{t - 1} + (J - M^s) \cdot \hat{x}'_{t - 1},
\end{align}
where $M^s$ is the mask for the source image, $J$ is an all-ones matrix of the same dimensions as $x_0$, and $\hat{x}_T \sim \mathcal{N}(0, I)$ is the initialized noise for the denoising process. The detailed visualization is shown in Appx~\ref{sec:algo}.

\subsection{Last-\texorpdfstring{$K$}{K} Step Denoising} \label{sec:last-k}
The DDIM~\cite{song2021denoising} or other approaches~\cite{lu2022dpm,zhao2024unipc} are often utilized to accelerate the denoising process by reducing the number of time steps. Despite its efficiency, this method may induce instability and diminish image quality in scenarios where models are trained on constrained or small-scale datasets. Consequently, a longer or full-time step is frequently required to achieve superior generation fidelity.

Unlike the original way, which starts with pure noise, we aim to modify the existing images, allowing us to bypass the initial sampling of pure noise. Inspired by \cite{meng2022sdedit}, we modify the initial target for denoising in Eq.~\ref{eq:recon} by adding noise from an intermediate time step $K$ to the image, simulating the noisy output $x_K$, and proceed to denoise the $x_K$ from $K - 1$ to 0 with DDPM. This allows our pipeline to achieve cross-domain translation without compromising speed.

\begin{figure}[t]
  \centering
  \includegraphics[width=0.9\linewidth]{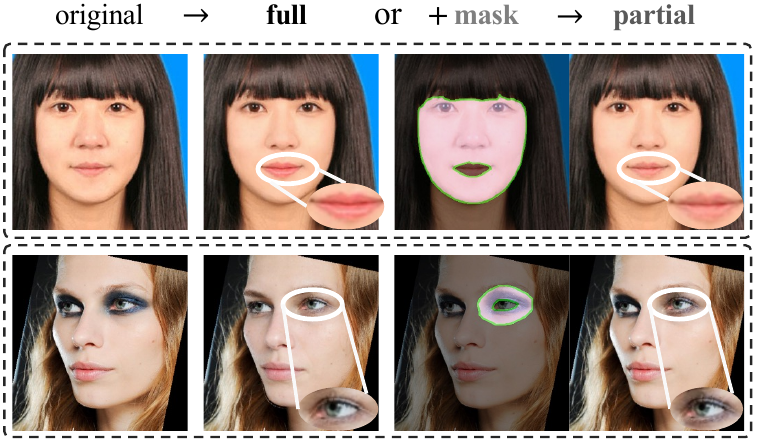}
  \caption{Beauty Filter (\textbf{top}) and Makeup Removal (\textbf{bottom}). We also provide the partial editing on the right.}
  \label{fig:domain-translation}
  \vspace{-10pt}
\end{figure}

\subsection{Makeup Tasks with MAD} \label{sec:makeup-tasks}

\paragraph{Beauty Filter and Makeup Removal.} To represent the makeup and non-makeup domains, we simply define two learnable embeddings. These embeddings are the conditional input to learn with the diffusion model to construct the image within the assigned domain. For beauty filter and makeup removal, we simply set $l_{so}$ to the makeup and non-makeup embedding, respectively, and $l_{ta}$ to the non-makeup and makeup embedding and perform a translation with our pipeline. In other words, by leveraging the model's adeptness in cross-domain translation, the learnable domain embeddings serve dual functions: enhancing non-makeup images through applying beauty filter and the precise removal of makeup from styled images. Fig.~\ref{fig:domain-translation} illustrates this dual capability, which manifests that our approach is able to remove complex makeup components like eye shadow and also allows for targeted beautification or removal.

\begin{figure*}[t]
  \centering
  \includegraphics[width=0.9\linewidth]{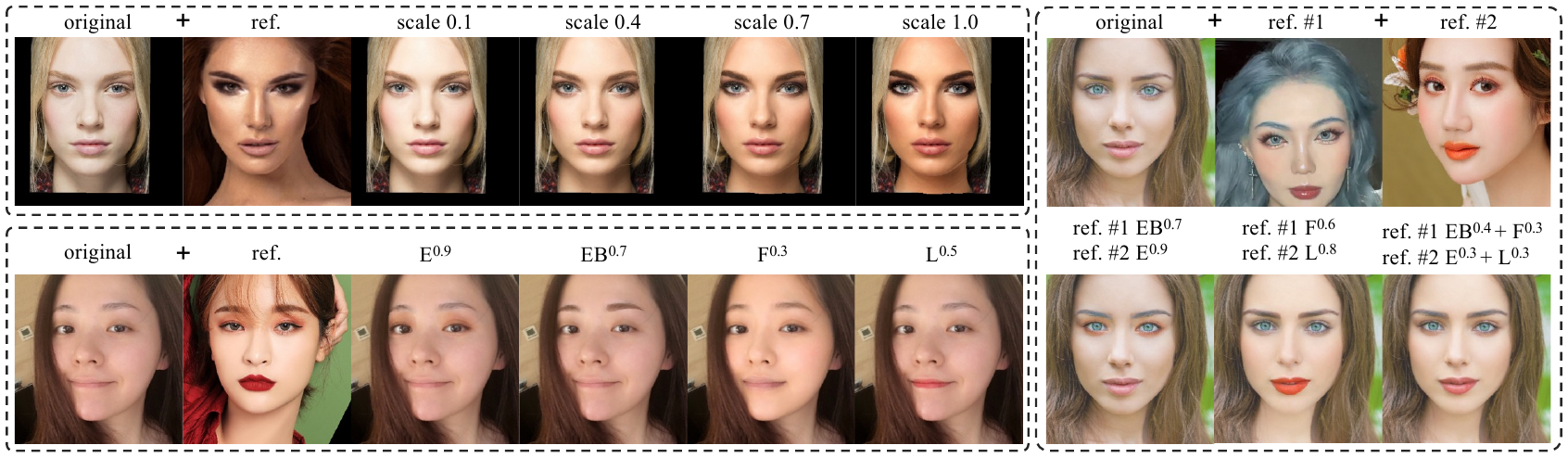}
  \caption{Makeup Transfer with different scales and styles. \textbf{Top Left}: Adjustable makeup scales. \textbf{Bottom Left}: Transfer intensities for individual components. \textbf{Right}: Combinations of scales and styles from different images for various components. Notations ``E'', ``EB'', ``F'', and ``L'' denote eyes, eyebrows, face, and lips, respectively, with superscript numbers indicating the component scale.}
  \label{fig:scale_comp_combine}
  \vspace{-5pt}
\end{figure*}

\begin{figure}[t]
  \centering
  \includegraphics[width=\linewidth]{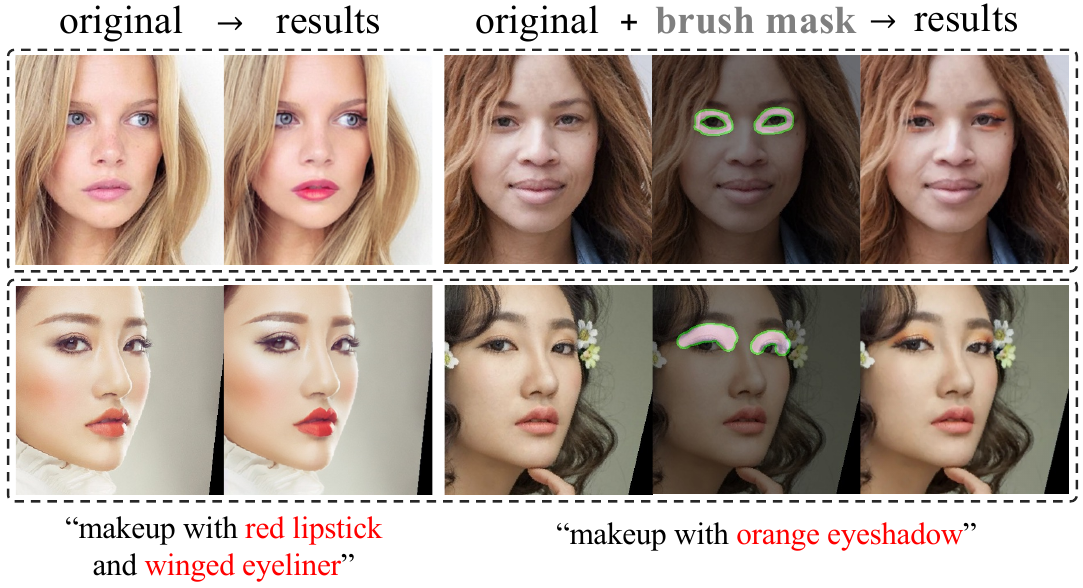}
  \caption{Text modification for non-makeup (\textbf{top}) and makeup (\textbf{bottom}) images.}
  \label{fig:text_modification}
  \vspace{-15pt}
\end{figure}

\paragraph{Text Modification.} Using textual descriptions as guidance allows flexible control without image reference. We adopt the text encoder from CLIP~\cite{radford2021learning} to extract the input text embeddings. The text encoder is fixed during diffusion training to maintain its powerful ability to provide text features. Afterward, we can alter the lip color, for example, with text by encoding the descriptor ``\textit{face without makeup}'' as $l_{so}$ and subsequently applying ``\textit{makeup with bold red lipstick}'' as $l_{ta}$ to achieve the desired modification. The same idea can also be applied to edit the makeup style from a makeup image with text for flexible usage. Fig.~\ref{fig:text_modification} showcases examples of our text-to-makeup capabilities with our without mask trained with the MT-Text dataset.\footnote{The details of the dataset are later discussed in Sec.~\ref{sec:text-data}.}

\paragraph{Makeup Transfer.} In the context of makeup transfer, our model is tasked with transforming an image to align with a reference style. Instead of using an external feature extractor, we aim to provide a simple training-free approach without additional modules. Adopting the last-$K$ step denoising, the initial target used for the generation still includes its original pattern since we only add noise to simulate the noisy output instead of sampling from pure noise. Consequently, we can embed the reference style directly into $x_0$ by blending it with the reference image. Hence, we consider the pure reference image to be the style embedding. To further consider the original input and output domain, the blended result and the makeup embedding act as $l_{ta}$ to help $x_0$ move toward the reference style within the makeup domain, and the original facial information encoded with the non-makeup embedding as $l_{so}$ prevents the output from moving far away from the original facial identity. As such, we can obtain the transfer result having the reference style with the original face appearance.

Specifically, we replace the original facial image $x_0$ with the blended image made from the source and the target image to directly embed the reference style. To obtain the blended image, we first compute a warping matrix derived from two sets of triangular meshes from the source and reference images and use the warping matrix to align the reference image with the source image. These meshes are generated using Delaunay triangulation based on facial landmarks, which can be acquired through standard detection approaches or a face mesh. Taking into account a source image $x_0$, a reference image $y_0$, and a designated warping function $F_{\text{warp}}$, the blended image $x'_0$ is given by:
\begin{equation}\label{eq:single-blend}
x'_0 = (J - \alpha) \cdot x_0 + \alpha \cdot F_{\text{warp}}(y_0),
\end{equation}
where $\alpha$ represents the weight matrix of the same dimensions as $x_0$, \textit{i.e.}, $[\alpha]_{ij} \in [0, 1]$ represents the element in the $i$-th row and $j$-th column of $\alpha$. By adjusting the blending weight, we can customize the intensity of different components. For example, assigning a higher $\alpha$ value to the eyes while setting a lower value to the lips allows accentuation of the eyes' makeup but reduces the style effect of lips. Fig.~\ref{fig:scale_comp_combine} demonstrates that the proposed MAD can perform makeup transfer with different scales, styles, and components.

\paragraph{Multi-Makeup Transfer.} We have refined the blending equation for makeup transfer involving multi-makeup and multi-scale across different reference images, as shown one the right of Fig.~\ref{fig:scale_comp_combine}. In a scenario involving a set of images $L$ targeted for makeup transfer, each is associated with a specific mask $M^l$ to choose one or more components, and a corresponding scale factor $\alpha^l$, the formulation for the blended image $x'_0$ is then refined as follows:
\begin{equation}\label{eq:multi-blend}
   x'_0 = \sum_{l \in L} (J - \alpha^l) \cdot \left(J - F_{\text{warp}}(M^l)\right) \cdot x_0 + \alpha^l \cdot F_{\text{warp}}(M^l \cdot y^l_0),
\end{equation}
where $\sum_{l \in L}\alpha^l = J$. This integrates various styles and scales from distinct reference images.

\paragraph{Component Asynchronous Masking.} Although using the blended target $x'_0$ facilitates the makeup style to the generation result, the components with smaller areas, such as the lips and eyebrows, may not effectively transfer the new makeup style. This downside is probably due to a heavy noise added to $x'_0$ as a larger denoising step is given, potentially diluting the impact of the blending style. The issue can be more severe in small regions with minimal information since our model has not been exposed to certain styles previously, and the model may misconstrue the style as noise and ignore the blended style. To address this issue, we introduce component asynchronous masking (CAM), where different denoising steps are assigned to different components. Although termed ``asynchronous,'' this method only varies the step $K$; the process itself can be executed simultaneously. We denote the source image mask $M^s$ with a subscript $t$ to indicate the mask used for the inpainting at the time step $t$. Specifically, $M^s_t$ for a set of components $C$ can be obtained as follows:
\begin{equation}
    M^s_t = \sum_{c \in C} \mathds{1}_{\{t > t_c\}} \cdot M^c,
\end{equation}
where $\mathds{1}_{\{ \}}$ is the indicator function, $t_c$ is the starting time for inpainting of component $c$, and $M^c$ is the mask for component $c$. This represents a component $c$ would have no mask to preserve its style when $t$ is smaller than $t_c$. We often set a smaller $t_c$ for smaller components to preserve its style until a smaller noise is added. In our settings, $C$ includes ``eyebrows'', ``eyes'', ``lips'', and ``face''. With CAM, we can obtain better transfer quality for non-seen styles or small facial areas. We provide the algorithm in Appx.~\ref{sec:algo}.

\begin{figure}[t]
  \centering
  \includegraphics[width=0.8\linewidth]{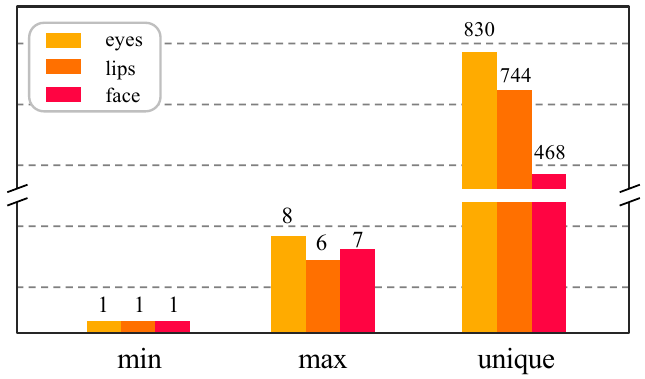}
  \caption{Statistics of the extended text data for the MT dataset. For better illustration, the scale for the ``unique'' y-axis is different.}
  \label{fig:statistics}
\end{figure}

\begin{figure}[t]
  \centering
  \includegraphics[width=\linewidth]{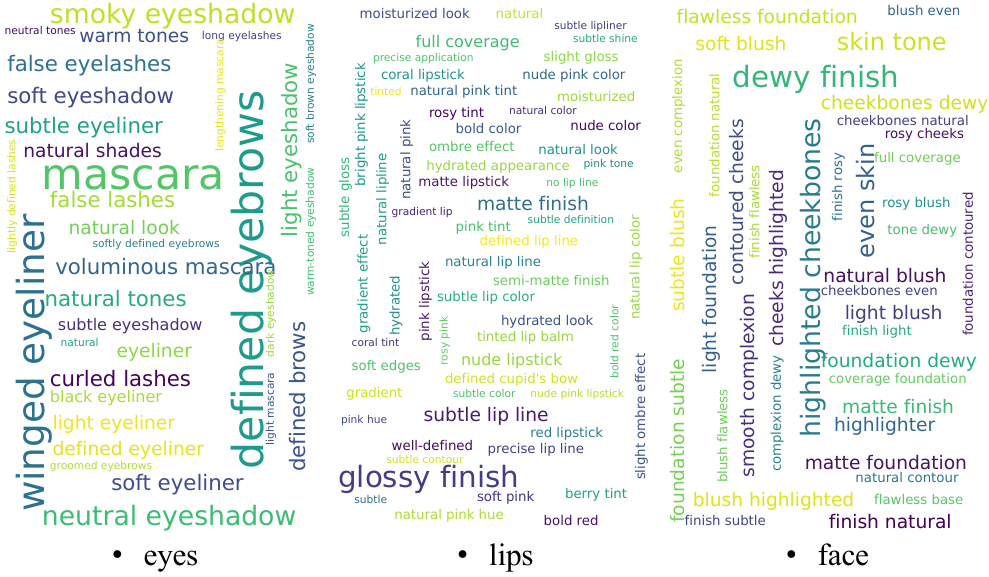}
  \caption{Word cloud of MT-Text for different areas.}
  \label{fig:word_cloud}
  \vspace{-10pt}
\end{figure}

\section{Dataset for Text-to-Makeup}\label{sec:text-data}

To advance research in text-to-makeup editing, we present MT-Text, an enriched version of the MT dataset~\cite{li2018beautygan} with textual annotations. This process, detailed in Appx.~\ref{sec:mt-text}, involves a zero-shot labeling process utilizing the GPT-4 Vision (GPT-4V) model~\cite{achiam2023gpt} to generate initial descriptions for three critical facial components: ``skin'', ``eyes'', and ``lips''. This minimizes the human effort required for labeling by providing a preliminary set of descriptive texts. Each generated description undergoes a meticulous review process to ensure its correctness. To ensure the quality of these annotations, the following criteria are implemented:

\begin{itemize}
    \item \textbf{Removing Redundancy}: Any repetitive terms are eliminated. Synonyms that have similar meanings are preserved to increase diversity. For instance, one of ``cat-eye'' and ``cat eye'' is removed, while both ``light eyeshadow'' and ``subtle eyeshadow'' are preserved.
    \item \textbf{Fixing Wrong Descriptions}: Corrections are made for any descriptions that misrepresent the makeup, such as confusing matte with glossy finishes, and we ensure that each facial component receives at least one label.
\end{itemize}

The MT-Text ultimately comprises 2,719 images, each meticulously annotated to reflect detailed and nuanced makeup styles. Fig.~\ref{fig:statistics} presents the dataset's statistics, with the columns labeled ``min'' and ``max'' indicating the variability in the number of descriptions per image component. This variability underscores our commitment to providing comprehensive annotations. The "unique" column highlights the richness of the dataset, showing the total number of distinct text descriptions crafted for various makeup styles. Each facial component is labeled with at least one descriptor, with some entities receiving up to eight style annotations. Moreover, 96\% of the entities are labeled with more than one descriptor, fostering a diverse set that can enhance the model's ability to understand and generate a wide range of makeup styles from textual descriptions. Fig.~\ref{fig:word_cloud} illustrates a word cloud for each component, visualizing the various terms used in our text annotations. Common terms include ``soft eyeliner'' for eyes, ``ombre effect'' for lips, and ``dewy finish'' for the face. These annotations enrich the dataset and simplify user interaction, making it more intuitive and user-friendly for real-world applications.

\section{Experiments}

\paragraph{Implementation.} Our model, abbreviated as \abb, is implemented in PyTorch and trained using a mixed precision setting on a Nvidia RTX 3090 GPU. In makeup transfer tasks, we set K to 180 for the face, 100 for the eyes, and 80 for the lips and eyebrows. We use the AdamW optimizer with a learning rate of $10^{-4}$, a weight decay of $10^{-2}$, and a batch size of 32. The AdamW hyperparameters, $(\beta_1, \beta_2)$, are set to $(0.95, 0.999)$. The training process spans 87,500 iterations with a 1000-step linear warm-up learning rate scheduler. For data augmentation, we apply only \textit{RandomHorizontalFlip} to transform the image.

\paragraph{Dataset.} Our evaluation is conducted using the MT dataset~\cite{li2018beautygan}, including 1,115 non-makeup and 2,719 makeup images. We also include the Wild~\cite{liu2021psgan++}, having 385 makeup images, and the BeautyFace (BF)~\cite{yan2023beautyrec}, consisting of 3000 makeup images, to examine the transfer ability for a diverse range of contemporary styles and head angles.

\paragraph{Evaluation Plan.} We assess the efficacy of our \textit{makeup removal} technique in Sec.~\ref{sec:exp-removal}, the effectiveness of our \textit{makeup transfer} methodology in Sec.~\ref{sec:exp_transfer}, and the capabilities of our \textit{text-to-makeup} application in Sec.~\ref{sec:exp-text}. Additionally, to analyze the impact and necessity of individual components, we conduct an ablation study in Sec.~\ref{sec:abla}.

\subsection{Makeup Removal}\label{sec:exp-removal}
To assess the model's ability to remove makeup, we examine the output of makeup transfer from Sec.~\ref{sec:exp_transfer} as the input for style removal. The original non-makeup images then serve as the reference style to guide the removal. For fairness, all methods are tested using the same set of inputs.

\paragraph{Metrics.} To evaluate the performance of makeup removal, we use the non-makeup images as the ground truth and measure the similarity between the original and the makeup-removed images using two metrics: Structural Similarity Index (SSIM)~\cite{wang2004image} and Peak Signal-to-Noise Ratio (PSNR). Higher values in both metrics indicate greater similarity between the compared images, reflecting more effective makeup removal.

\paragraph{Results.}  We compare the makeup removal performance with state-of-the-art approaches in Tab.~\ref{tab:removal}. Our approach features two removal methods: direct embedding removal, denoted with a $^\dag$, and reference style-based removal, marked with a $^\ddag$. By removing the makeup with the non-makeup embedding, our approach achieves the second-best results on SSIM and surpasses most on PSNR. Furthermore, by removing makeup with the reference style, our approach obtains the best results on both metrics as it gets more clues of the original appearance compared with using only the embedding. Visual comparison is provided in Appx.~\ref{sec:visual-comp}

\begin{table}[t]
  \caption{Results of the makeup removal on the MT and BF dataset. \textbf{Bold} indicates the best and \underline{underline} denotes the second best.}
  \label{tab:removal}
  \centering
  \resizebox{0.9\linewidth}{!}{\begin{tabular}{l c c c c}
    \toprule
    \multirow{2.5}{*}{Method} & \multicolumn{2}{c}{MT} & \multicolumn{2}{c}{BF}\rule{0pt}{2.5ex}\rule[-1ex]{0pt}{0pt} \\ \cmidrule(lr){2-3}  \cmidrule(lr){4-5}
    & SSIM $\uparrow$ & PSNR $\uparrow$ & SSIM $\uparrow$ & PSNR $\uparrow$ \rule{0pt}{2.5ex}\rule[-1ex]{0pt}{0pt} \\
    \midrule
    PSGAN++~\cite{jiang2020psgan} & 0.739 & 21.837 & 0.756 & 22.419\rule{0pt}{2.5ex}\rule[-1ex]{0pt}{0pt}\\
    SCGAN~\cite{deng2021spatially} & 0.885 & 25.798 & 0.889 & \underline{26.290}\rule{0pt}{2.5ex}\rule[-1ex]{0pt}{0pt}\\
    CPM~\cite{nguyen2021lipstick} & 0.918 & \underline{26.728} & 0.918 & 26.728\rule{0pt}{2.5ex}\rule[-1ex]{0pt}{0pt} \\
    ELeGANt~\cite{yang2022elegant} & 0.749 & 23.256 & 0.766 & 24.032\rule{0pt}{2.5ex}\rule[-1ex]{0pt}{0pt}\\
    SSAT~\cite{sun2022ssat} & 0.886 & 24.743 & 0.887 & 24.891\rule{0pt}{2.5ex}\rule[-1ex]{0pt}{0pt}\\
    BeautyREC~\cite{yan2023beautyrec} & 0.868 & 25.189 & 0.871 & 25.593\rule{0pt}{2.5ex}\rule[-1.6ex]{0pt}{0pt} \\
    \midrule
    MAD$^\dag$ (ours) & \underline{0.922} & 25.304 & \underline{0.923} & 25.658\rule{0pt}{2.8ex}\rule[-1ex]{0pt}{0pt} \\
    MAD$^\ddag$ (ours) & \textbf{0.936} & \textbf{28.263} & \textbf{0.936} & \textbf{28.538}\rule{0pt}{2.5ex}\rule[-1ex]{0pt}{0pt} \\
  \bottomrule
  \end{tabular}}
\end{table}

\begin{table*}[t]
  \caption{Results of the makeup transfer on MT, Wild, and BF dataset with identity score (IS), precision (Pr), recall (Rc), KID, and user study. \textbf{Bold} indicates the best and \underline{underline} denotes the second best.}
  \label{tab:eval}
  \centering
  \resizebox{=0.95\linewidth}{!}{\begin{tabular}{l c c c c c c c c c c c c c c c}
    \toprule
    \multirow{2}{*}{Method} & \multicolumn{4}{c}{MT} & \multicolumn{4}{c}{Wild} & \multicolumn{4}{c}{BF} & \multicolumn{3}{c}{User Study Top-1} \rule{0pt}{2.5ex}\rule[-1ex]{0pt}{0pt} \\ \cmidrule(lr){2-5} \cmidrule(lr){6-9} \cmidrule(lr){10-13} \cmidrule(lr){14-16}
    & IDS $\uparrow$ & Pr $\uparrow$ & Rc $\uparrow$ & KID $\downarrow$ & IDS $\uparrow$ & Pr$\uparrow$ & Rc $\uparrow$ & KID $\downarrow$ & IDS $\uparrow$ & Pr $\uparrow$ & Rc $\uparrow$ & KID $\downarrow$ & MT & BF & Avg $\uparrow$ \rule{0pt}{2.5ex}\rule[-1ex]{0pt}{0pt} \\
    \midrule
    PSGAN++~\cite{liu2021psgan++} & 0.314 & 0.916 & 0.025 & \underline{0.039} & 0.336 & 0.923 & 0.140 & 0.024 & 0.842 & 0.906 & 0.064 & \underline{0.029} & 0\% & 0\% & 0\% \rule{0pt}{2.5ex}\rule[-1ex]{0pt}{0pt}\\
    SCGAN~\cite{deng2021spatially} & 0.219 & \underline{0.925} & 0.050 & \underline{0.039} & 0.082 & 0.874 & 0.045 & 0.110 & 0.213 & 0.860 & 0.068 & 0.048 & 0\% & 0\% & 0\% \rule{0pt}{2.5ex}\rule[-1ex]{0pt}{0pt}\\
    CPM~\cite{nguyen2021lipstick}  & \underline{0.914} & 0.851 & 0.009 & 0.043 & \underline{0.873} & \underline{0.937} & \underline{0.250} & \underline{0.018} & \underline{0.894} & \underline{0.915} & \textbf{0.327} & 0.049 & 10\% & 0\% & 5\% \rule{0pt}{2.5ex}\rule[-1ex]{0pt}{0pt} \\
    EleGANt~\cite{yang2022elegant}  & 0.203 & 0.851 & 0.009 & 0.043 & 0.233 & 0.885 & 0.077 & 0.021 & 0.208 & 0.815 & 0.067 & 0.042 & \underline{30\%} & \underline{30\%} & \underline{30\%} \rule{0pt}{2.5ex}\rule[-1ex]{0pt}{0pt} \\
    SSAT~\cite{sun2022ssat}  & 0.283 & 0.869 & 0.004 & 0.084 & 0.259 & 0.897 & 0.097 & 0.030 & 0.251 & 0.851 & 0.114 & 0.069 & 10\% & 0\% & 5\% \rule{0pt}{2.5ex}\rule[-1ex]{0pt}{0pt}\\
    BeautyREC~\cite{yan2023beautyrec}  & 0.184 & 0.867 & 0.008 & 0.066 & 0.123 & 0.900 & 0.063 & 0.060 & 0.184 & 0.886 & 0.156 & 0.032 & 10\% & 0\% & 5\% \rule{0pt}{2.5ex}\rule[-1ex]{0pt}{0pt}\\
    \abb (Ours) & \textbf{0.933} & \textbf{0.948} & \textbf{0.106} & \textbf{0.018} & \textbf{0.912} & \textbf{0.958} & \textbf{0.329} & \textbf{0.003} & \textbf{0.935} & \textbf{0.932} & \underline{0.317} & \textbf{0.025} & \textbf{40\%} & \textbf{70\%} & \textbf{55\%} \rule{0pt}{2.5ex}\rule[-1ex]{0pt}{0pt}\\
  \bottomrule
  \end{tabular}}
\end{table*}

\subsection{Makeup Transfer}\label{sec:exp_transfer}

For evaluation, we use 1,109\footnote{We remove six non-makeup images as these images have no eyes.} non-makeup images and randomly selected an equivalent number of unique makeup images as references from the MT and BF datasets. Since the Wild dataset only has 385 makeup images, we randomly selected the same number of non-makeup images.

\paragraph{\textbf{Metrics}} We follow the popular metrics for image generation with Precision, Recall~\cite{sajjadi2018assessing}, and KID\footnote{FID can cause a larger bias with only a thousand images.}~\cite{bińkowski2018demystifying}. However, directly comparing the distribution cannot be accurate, as it could be challenging to separate the identity quality and the makeup style, and there is a potential where models could opt \textbf{NOT} to perform transfer and copy the original image to maximize these scores. Consequently, we perform feature manipulation to compare the feature distribution and adopt a finetuned ResNet-50~\cite{he2016deep} to evaluate identity preservation. We define the non-makeup set, makeup set, transfer results, and feature extractor as $S$, $R$, $S_R$, and $F$, respectively.

\begin{itemize}
    \item \textbf{Precision}: The identity quality is evaluated through the precision of $F(S_R) - F(R)$ against $F(S)$, with the removal of the makeup to reveal the original facial feature.
    \item \textbf{Recall and KID}: For makeup style, the recall and KID are calculated between $F(S_R) - F(S)$ and $F(R) - F(T(R))$, leaving the makeup feature by removing the base identity, where $T(\cdot)$ is the makeup removal transformation provided by our approach\footnote{The weight is \textbf{different} from makeup transfer for fairness.}.
    \item \textbf{Identity Score (IDS)}: The IDS is evaluated by averaging the probability of the true identities obtained from the ResNet model. We finetune the ResNet model to classify the identity with all the images within all three datasets.
\end{itemize} 

\paragraph{User Study.} We conduct a user study by randomly selecting 20 non-makeup images and randomly choosing 10 reference images from the MT dataset and another 10 from the BF dataset. We collect 30 responses, in which participants are instructed to assess the generated samples across three primary dimensions, ordered by importance: visual quality, transfer quality, and preservation quality. The evaluation uses a ``winner-takes-all'' (top-1) to assess the results.

\paragraph{Results.} The evaluation results, detailed in Tab.~\ref{tab:eval}, highlight our method's superiority on MT, Wild, and BF datasets, with only getting a lower score on Pr in the BF dataset. This success is attributed to our model's capability to generate high-quality images and preserve the subject's details throughout the encoding process, thereby ensuring exceptional identity preservation and visual quality, as evidenced by the IDS and Pr metrics. Furthermore, our method's blending technique allows for the explicit integration of makeup styles directly onto the image, facilitating accurate makeup transfer as validated by the recall and KID metrics. Furthermore, the user study results further underscore our dominance with the top-1 score in both datasets.

\subsection{Text-to-Makeup}\label{sec:exp-text}

\begin{table}[tb!]
  \caption{Comparison of text-to-makeup editing. $^*$ represents the finetuned version of the MT-Text dataset.}
  \label{tab:text-editing}
  \centering
  \resizebox{0.85\linewidth}{!}{\begin{tabular}{l c c c c}
    \toprule
     Method & $\text{CLIP}_{T} \uparrow$ & $\text{CLIP}_{I} \uparrow$ & $\text{CLIP}_{S }\uparrow$\rule{0pt}{2.5ex}\rule[-1ex]{0pt}{0pt}\\
    \midrule
    SD Inpainting~\cite{rombach2022high} & 0.292 & 0.617 & 0.598\rule{0pt}{2.5ex}\rule[-1ex]{0pt}{0pt}\\
    Null Inversion~\cite{mokady2023null} & \textbf{0.300} & 0.594 & 0.620\rule{0pt}{2ex}\rule[-1ex]{0pt}{0pt} \\
    CycleDiffusion~\cite{wu2023latent} & 0.281 & 0.695 & 0.611\rule{0pt}{2ex}\rule[-1ex]{0pt}{0pt}\\
    Prompt2Prompt~\cite{hertz2023prompttoprompt} & 0.284 & 0.728 & 0.641\rule{0pt}{2ex}\rule[-1.6ex]{0pt}{0pt}\\
    \midrule
    MAD\phantom{$^*$} (ours) & 0.260 & 0.916 & 0.655\rule{0pt}{2.4ex}\rule[-1ex]{0pt}{0pt}\\
    MAD$^*$ (ours) & 0.266 & \textbf{0.931} & \textbf{0.662}\rule{0pt}{2ex}\rule[-1ex]{0pt}{0pt}\\
  \bottomrule
  \end{tabular}}
\end{table}

To evaluate the performance of text-to-makeup, we randomly selected 100 non-makeup images and 100 makeup images with associated text annotations from the MT-Text.

\paragraph{Metrics.} We employ metrics based on CLIP~\cite{radford2021learning}. Specifically, we utilize $\text{CLIP}_{T}$ to evaluate the alignment between the text descriptions and the makeup results, $\text{CLIP}_{I}$ for identity preservation, and $\text{CLIP}_{S}$ for the similarity between the output features and the reference makeup features.

\paragraph{\textbf{Results}} We compare our approach with the Stable Diffusion (SD) Inpainting~\cite{rombach2022high}, Null-Text Inversion~\cite{mokady2023null}, CycleDiffusion~\cite{wu2023latent}, and Prompt2Prompt~\cite{hertz2023prompttoprompt} in Tab.~\ref{tab:text-editing}. For fairness, we adopt the same pre-trained stable diffusion weights in this experiment and set the guidance scale to 15 for all the approaches. For the $\text{CLIP}_{T}$ score, other approaches obtain a better score as they tend to provide a more extreme style but ignore the original identity. This can be observed by $\text{CLIP}_{I}$ metrics where we outperform other approaches in a larger gap. As for $\text{CLIP}_{S}$, we achieve the highest score. These results show that our approach not only preserves the identity in a better way but also provides the correct makeup style from the text. We finetune our model and indicate this with a $^*$ in the results to highlight the benefits of our tailored text annotation dataset. Both text, identity, and style scores show improvements in finetuning with the MT-Text, showcasing its effectiveness. The visual comparison is provided in Appx.~\ref{sec:visual-comp}.

\subsection{Ablation Study}\label{sec:abla}
We perform ablation studies on the MT and BF datasets with makeup transfer to evaluate the design.

\paragraph{Last-$K$ vs. DDIM.} Analysis from the first two rows of the Tab.~\ref{tab:abla} reveals that DDIM can deteriorate the visual and stylistic quality of makeup transfer since we only have few training data to train the model, which is challenging to approximate the correct trajectory with a faster sampling approach. Thus, we deduce that using the last-$K$ method can accelerate the generation process without sacrificing image quality, starting the transfer process with the clue of the original face and adding target styles to it.

\paragraph{Component Asynchronous Masking (CAM).}
Our ablation study of CAM, presented in the last two rows of Tab.~\ref{tab:abla}, shows a slight improvement in image quality and minimal differences in style coverage. Though quantitative measures show only modest improvements, the visualization in Appx.~\ref{sec:visual-comp} shows noticeable improvement with CAM.

\begin{table}[tb!]
  \caption{Ablation study of component asynchronous masking (CAM) and last-$K$ step denoising (last-$K$).}
  \label{tab:abla}
  \centering
  \resizebox{\linewidth}{!}{\begin{tabular}{c c c c c c c c c}
    \toprule
     \multirow{2}{*}{DDIM} & \multirow{2}{*}{CAM} & \multirow{2}{*}{last-$K$} & \multicolumn{2}{c}{MT} & \multicolumn{2}{c}{Wild} & \multicolumn{2}{c}{BF}\rule{0pt}{2.5ex}\rule[-1ex]{0pt}{0pt}\\ \cmidrule(lr){4-5} \cmidrule(lr){6-7} \cmidrule(lr){8-9}
     & & & Pr $\uparrow$ & KID $\downarrow$ & Pr $\uparrow$ & KID $\downarrow$ & Pr $\uparrow$ & KID $\downarrow$\rule{0pt}{2.5ex}\rule[-1ex]{0pt}{0pt} \\
    \midrule
    \checkmark  & & & 0.267 & 0.421 & 0.362 & 0.537 & 0.245 & 0.471\rule{0pt}{2.5ex}\rule[-1ex]{0pt}{0pt} \\
    \checkmark & \checkmark & & 0.234 & 0.767 & 0.359 & 0.768 & 0.237 & 0.738\rule{0pt}{2.5ex}\rule[-1.6ex]{0pt}{0pt} \\
    \midrule
    & \checkmark & & 0.947 & 0.019 & 0.958 & 0.003 & 0.927 & 0.026\rule{0pt}{2.8ex}\rule[-1ex]{0pt}{0pt}  \\
    & \checkmark & \checkmark & \textbf{0.948} & \textbf{0.018} & \textbf{0.942} & 0.003 & \textbf{0.937} & \textbf{0.025}\rule{0pt}{2.5ex}\rule[-1ex]{0pt}{0pt}\\
  \bottomrule
  \end{tabular}}
\end{table}
\section{Conclusion and Future Work}

This work introduces a cross-domain diffusion model with domain embeddings to unify various makeup applications into domain translation problems. Our approach translates images between domains without sampling from the random noise, allowing us to apply the last-$K$ step denoising to accelerate the generation process and preserve original facial information. Additionally, utilizing inpainting and component asynchronous masking, our model offers flexible partial area translation across different tasks and provides better makeup styles to small areas such as lips and eyebrows. To further advance text-to-makeup research and enhance practical utility, we provide the MT-Text dataset for text-to-makeup applications. In the future, we plan to increase the diversity of text annotations by including more contemporary or extreme styles. Furthermore, as existing studies primarily focus on 2D modeling, we plan to extend the domain to 3D space, facilitating transformations for 3D makeup applications for the anime or game industry.

{
    \small
    \bibliographystyle{ieeenat_fullname}
    \bibliography{main}
}

\begin{appendices}
    \appendix
    \clearpage
\setcounter{page}{1}

\twocolumn[{%
\renewcommand\twocolumn[1][]{#1}%
\maketitlesupplementary
\begin{center}
    \centering
    \captionsetup{type=figure}
    \includegraphics[width=\linewidth]{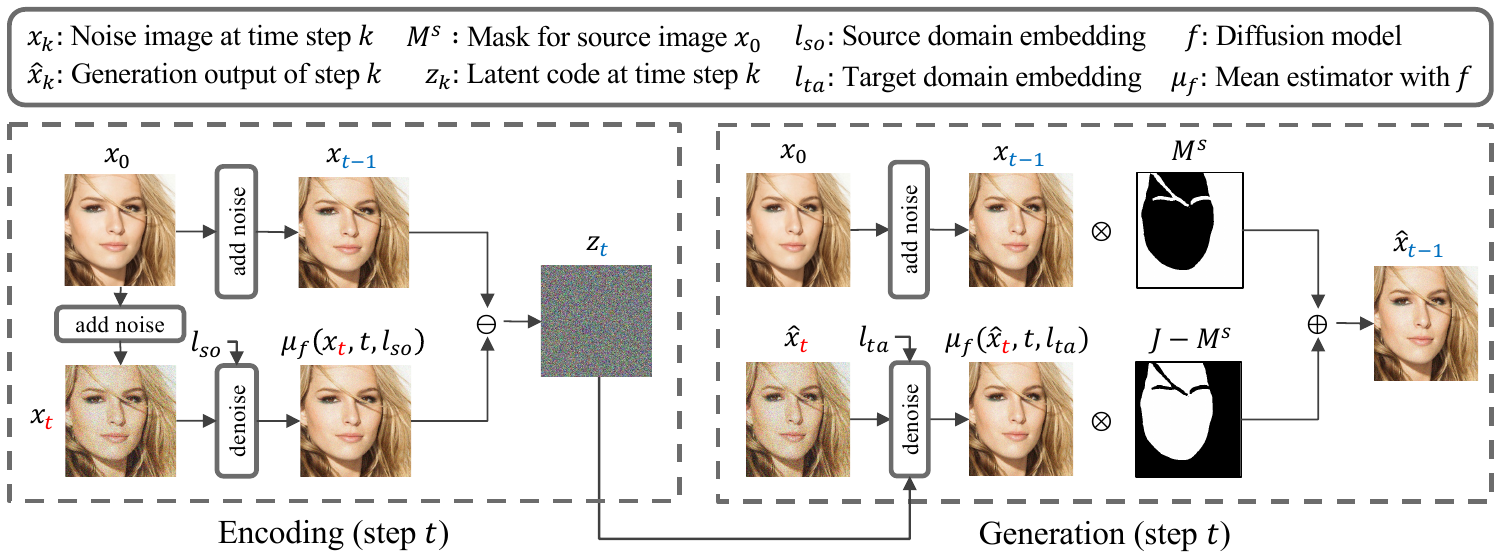}
    \captionof{figure}{Illustration of the cross-domain diffusion pipeline for time step $t$. Initially, the pipeline generates a latent code representing the source domain, which is subsequently used in the target domain generation to ensure detail preservation. During the generation phase, a preserved mask can be applied to maintain non-facial regions or to modify specific components.}
    \label{fig:pipe}
\end{center}%
}]

\section{Details of Makeup Transfer} \label{sec:algo}

Algorithm~\ref{alg:cross-domain-makeup-transfer} outlines the procedure for single makeup transfer. For multi-makeup transfer, we replace the blending equation in Eq.~\ref{eq:single-blend} with Eq.~\ref{eq:multi-blend}. Furthermore, Fig.~\ref{fig:pipe} offers a comprehensive visualization of the cross-domain translation process within our framework, as described in Sec.~\ref{sec:cross-domain-diffusion}.

\begin{algorithm}[ht]
  \caption{Makeup Transfer}
  \label{alg:cross-domain-makeup-transfer}
  \KwIn{step $K$, component set $C$, component masks $\{M^c | c \in C\}$ and starting time $\{t_c | c \in C\}$, source image $x_0$ and mask $M^s$, reference image $y_0$, model $f$, source and target embedding $l_{so}$ and $l_{ta}$, and scale $\alpha$}
  \KwOut{denoised output $\hat{x}_0$}
  Obtain source and reference facial mesh $R_s$ and $R_r$ \;
  Compute warping function $F_{\text{warp}}$ from $R_s$ and $R_r$ \;
  $x'_0 \leftarrow (J - \alpha) x_0 + \alpha F_{\text{warp}}(y_0)$ \tcp*{Perform Blending}
  $\hat{x}_K \sim q(\hat{x}_K | x_0')$ \tcp*{Obtain noisy input}
  \For{$t\leftarrow K$ \KwTo $1$}
  {
    \tcc{Obtain source facial latent code}
    $x_{t - 1} \sim q(x_{t - 1} | x_0) \ ; \ z_t = \frac{(x_{t - 1} - \mu_{f}(x_t, t, l_{so}))}{ \sigma_t}$ \;
    \tcc{Component Asynchronous Masking}
    $M^s_t = \sum_{c \in C} \mathds{1}_{\{t > t_c\}} \cdot M^c$ \;
    \tcc{Perform masking for preservation}
    $\hat{x}'_{t - 1} \leftarrow \mu_f(\hat{x}_t, t, l_{ta}) + \sigma_t z_t$ \;
    $\hat{x}_{t - 1} \leftarrow M^s_t \cdot x_{t - 1} + (J - M^s_t) \cdot \hat{x}'_{t - 1}$\;
  }
  \textbf{return} $\hat{x}_0$ \;
\end{algorithm}

\section{Annotation of MT-Text Dataset} \label{sec:mt-text}

\begin{figure*}[!ht]
  \centering
  \includegraphics[width=\linewidth]{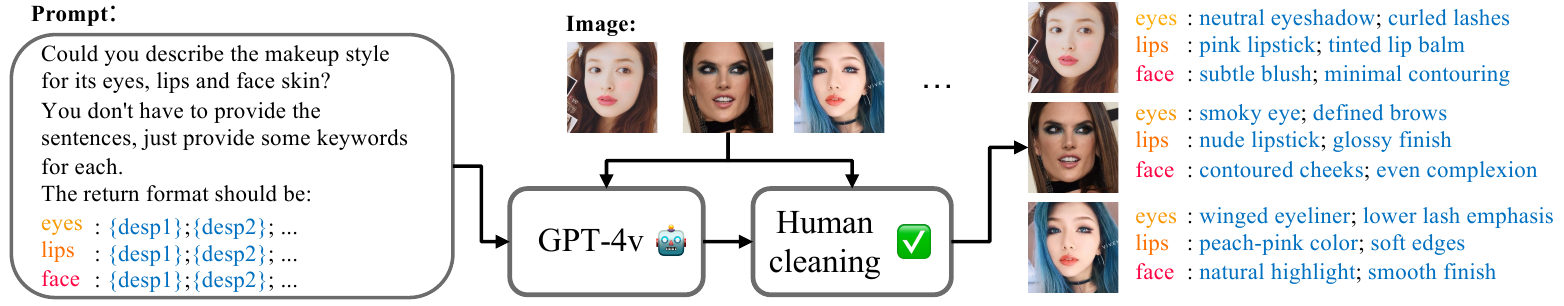}
  \caption{Illustration of our text labeling process. We first apply the GPT-4V model to provide a rough description of each area and clean the output to provide the correct labeling results.}
  \label{fig:text-labeling}
\end{figure*}

The detailed annotation process for the MT-Text dataset is illustrated in Fig.~\ref{fig:text-labeling}. Using prompt and input images, the GPT-4v agent generates an initial labeling for three facial regions: ``eyes,'' ``lips,'' and ``face,'' as depicted on the right.

\section{Visual Comparison}\label{sec:visual-comp}

\begin{figure*}[ht]
  \centering
  \includegraphics[width=\linewidth]{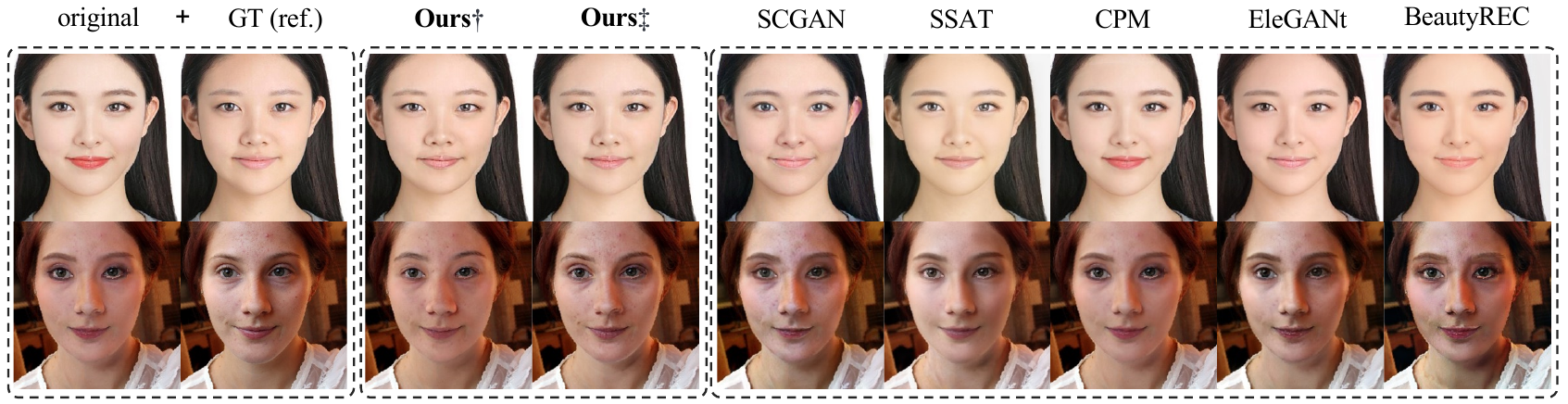}
  \caption{Visual comparison for makeup removal. The reference style of the first row is from the MT dataset, and the second is from the BF dataset. $\dag$ represents makeup removal with an embedding, and $\dagger$ represents the makeup removal with the reference style.}
  \label{fig:removal-comp}
  \vspace{-10pt}
\end{figure*}

\paragraph{Makeup Removal.} A visual comparison of makeup removal is provided in Fig.~\ref{fig:removal-comp}. Our method, using only the removal embedding, offers a good approximation of the subject's original appearance. However, incorporating the reference style significantly enhances the accuracy of the results. This improvement is quantifiable, as reflected by higher PSNR values when using the reference style compared to the embedding alone. Using only the embedding may not consistently align with the original features. Notably, when compared to other methods that also use reference styles, our approach achieves more precise removal of makeup on eyebrows, lips, and skin color.

\begin{figure}[t]
  \centering
  \includegraphics[width=\linewidth]{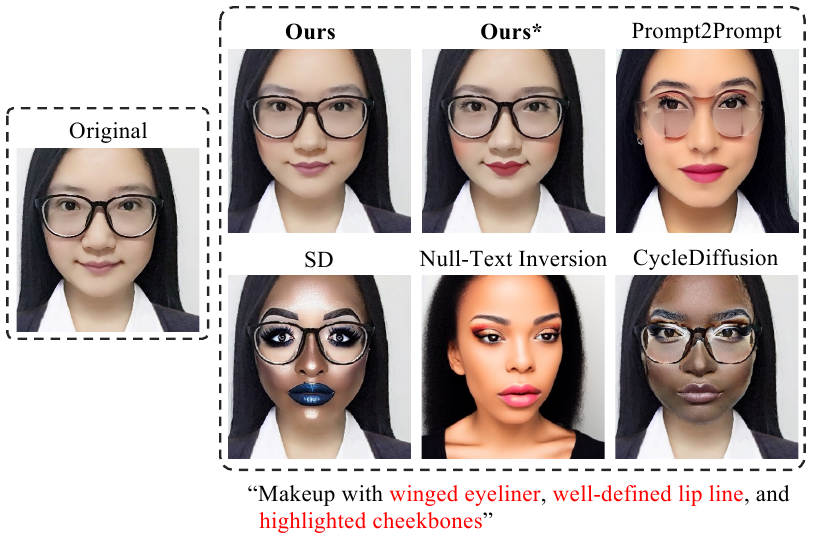}
  \caption{Visual comparison for text editing with the prompt at the bottom. $^*$ represents the finetuned version with MT-Text.}
  \label{fig:text-visual-comp}
\end{figure}

\paragraph{Text-to-Makeup.}  We provide the visual comparison in Fig.~\ref{fig:text-visual-comp}. Our approach can obviously preserve the identity better and still provide the correct makeup style. Other approaches tend to fit only the prompt without considering the original appearance. Compared with the non-finetuned version, the finetuned model can provide a more explicit style, such as a better illustration of Cupid’s Bow (with a defined lip line), highlighted cheekbones with brighter color, and using eyelashes to accentuate the winged eyeliner. 

\begin{figure*}[ht]
  \centering
  \includegraphics[width=\linewidth]{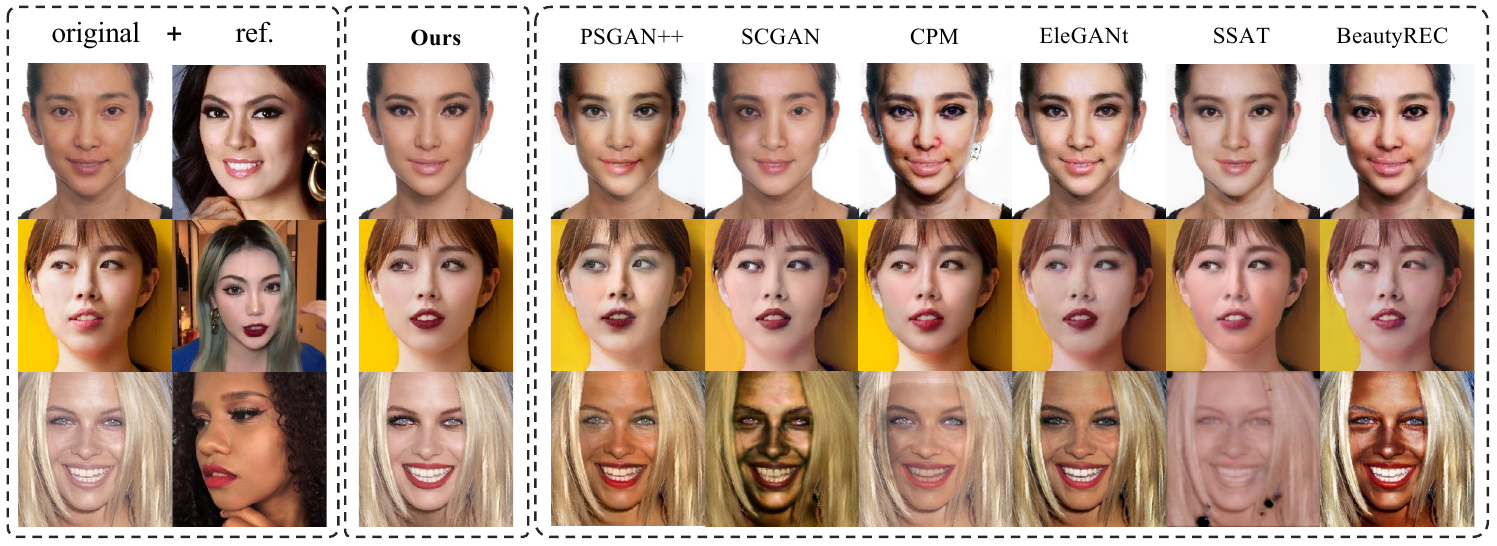}
  \caption{Visual comparison for makeup transfer. The reference image of the first row is from the MT dataset, the second row from the BF dataset, and the third row from the Wild dataset.}
  \label{fig:comp}
\end{figure*}

\paragraph{Makeup Transfer.} In Fig.~\ref{fig:comp}, we illustrate the effectiveness of our makeup transfer method in various scenarios. Our blending approach selectively transfers makeup styles without incorporating undesired elements, such as dark tones (first row), from the reference images, showcasing our method's precision in capturing and applying only relevant styles. Additionally, our method is not severely affected by new skin tone (third row) due to utilizing original encoding information during generation. Our approach also enables adaptability to slightly different head sizes (second row) and different poses (third row) due to utilizing warping for accurate alignment of makeup elements before generation, ensuring consistency across various facial orientations.

\begin{figure}[t]
  \centering
  \includegraphics[width=\linewidth]{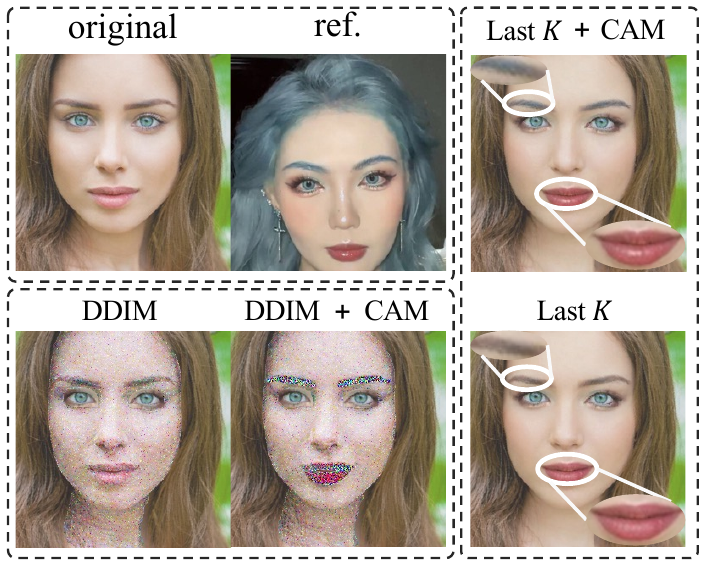}
  \caption{Visual comparison for ablation study. This figure contrasts the DDIM with our last-$K$ step approach and shows the impact of component asynchronous masking (CAM) on the results.}
  \label{fig:abla}
\end{figure}

\begin{figure}[t]
  \centering
  \includegraphics[width=\linewidth]{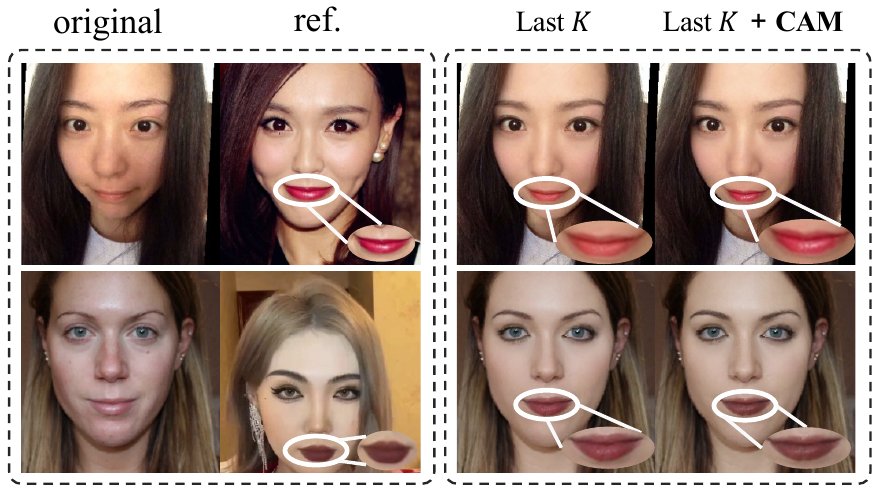}
  \caption{More examples to demonstrate that with component asynchronous masking (CAM), we can better transfer the styles, including luminance and color, for small areas.}
  \label{fig:cam-more-comp}
\end{figure}

\paragraph{Ablation Study.} A comparative visual demonstration for last-$K$ vs. DDIM is provided in Fig.~\ref{fig:abla} as evidence. Additionally, visual comparisons in Fig.~\ref{fig:abla} highlight CAM’s ability to capture finer details for small areas, thereby enhancing the fidelity and precision of the makeup representation. Additional examples in Fig.~\ref{fig:cam-more-comp} further demonstrate CAM’s effectiveness in dealing with small areas to improve the realism and precision of makeup application.

\begin{figure}[t]
  \centering
  \includegraphics[width=\linewidth]{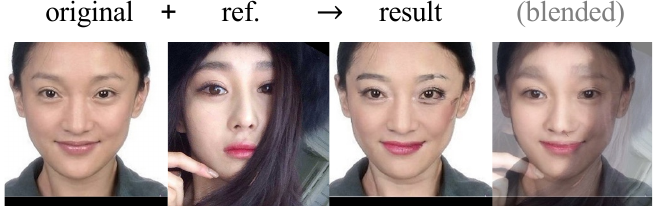}
  \caption{Illustration of the failure case due to facial occlusion. We provide the ``blended'' output on the rightmost to illustrate how the hair occlusion affects the transferred results.}
  \label{fig:occlusion}
\end{figure}

\section{Analysis of Last-K Step Denoising}

\begin{figure}[t]
  \centering
  \includegraphics[width=\linewidth]{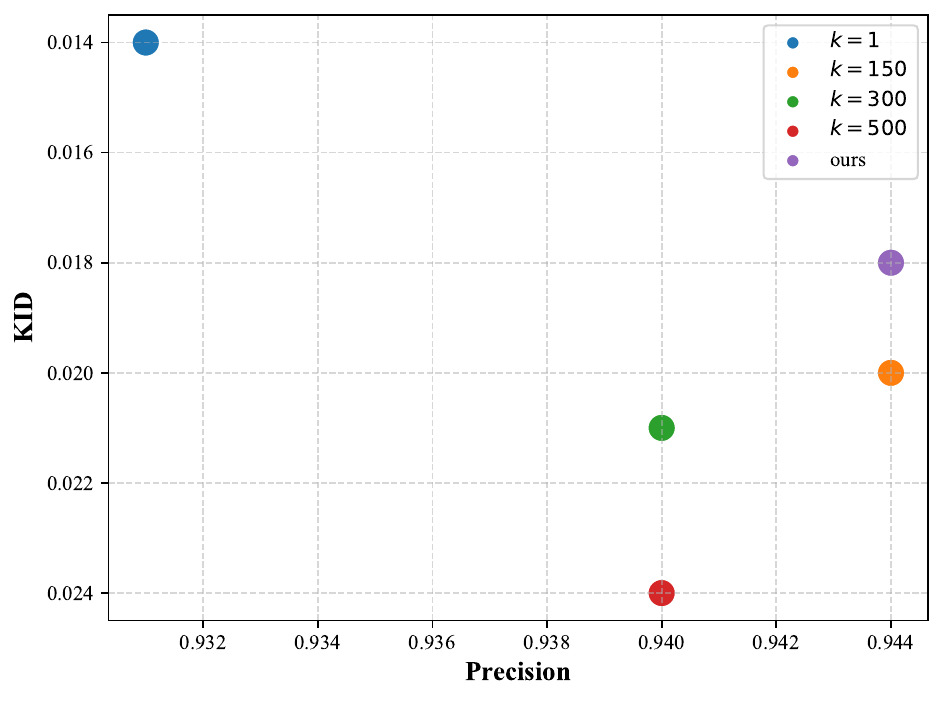}
  \caption{Illustration of different $k$ for last-step $k$.}
  \label{fig:analysisk}
\end{figure}

We examine the results across different values of $K$ and compare them with our component, asynchronous masking, as illustrated in Fig.~\ref{fig:analysisk}. As $K$ increases, style matching performance decreases, as indicated by higher KID scores. In contrast, larger values of $K$ yield better identity preservation. Among all methods, ours achieves the best identity quality while maintaining competitive style matching.

\section{Limitation and Failure Analysis}

Here, we discuss the limitations and present failure cases of our model, along with possible reasons for these failures. This analysis is intended to guide future research directions in the makeup field based on our work.

\paragraph{\textbf{Slow Generation Speed}} A notable challenge associated with our diffusion model is its relatively slow generation speed, with each image taking approximately 20 seconds to process, attributed primarily to the diffusion model's complexity. Although the process is optimized by focusing on the last $K$ steps, the method still requires hundreds of steps to achieve the desired results. Future improvements could aim at improving the generation speed, potentially through the adoption of a latent diffusion model approach, as suggested by~\cite{rombach2022high}, or employing our idea with the Rectified Flow~\cite{liu2023flow} for cross-domain translation.

\paragraph{\textbf{Facial Occlusion Challenges.}} Facial alignment, a critical step in our makeup transfer approach, can be problematic when faces are obscured by hair or other accessories. The style or the color for the occlusion part can be inconsistent with other areas, making the transfer process fail, as shown in Fig.~\ref{fig:occlusion}. Since the left face of the reference image is blocked by its hair, we can observe some artifacts on the left face of the transferred result. A promising area for future research based on our work could be mixing in feature space instead of directly applying blending for pure images. This approach has the potential to eliminate visible artifacts, thereby preventing inconsistent or weird results.

\paragraph{\textbf{Unnatural Text-to-Makeup Styles}} Although our approach supports text-to-makeup generation, it may produce unnatural makeup styles, even when accurately following the prompts. An example of this can be observed in the orange eyeshadow shown in Fig.~\ref{fig:text_modification}. To address this issue, we plan to collect more contemporary makeup data and incorporate novel or bold makeup styles to achieve natural and creative results.

\end{appendices}

\end{document}